\documentclass[twocolumn,a4paper]{article} %<--- For `pdflatex`' user

\usepackage{nolta2024}
\usepackage{amsmath} %<--- Please uncomment this command if you use amsmath package.
\usepackage{txfonts}
\usepackage{color}
\usepackage{graphics}
\usepackage{graphicx}
\usepackage{multirow}
\usepackage{hyperref}
\usepackage{placeins}
\usepackage{xspace}
\providecommand{\tabularnewline}{\\}

\newcommand\correspondingauthor{\thanks{Corresponding author: Thien Van Luong.}}

\begin{document}

\title{Effective Intrusion Detection for UAV Communications using Autoencoder-based Feature Extraction and Machine Learning Approach}

\author{
  {Tuan-Cuong Vuong${}^\dag$, Cong Chi Nguyen${}^\dag$, Van-Cuong Pham${}^\dag$,} 
   {
  Thi-Thanh-Huyen Le${}^*$, Xuan-Nam Tran${}^*$, and  Thien Van Luong${}^\ddag$\correspondingauthor}}
  
% \author{
%   Thi-Thanh-Huyen Le${}^*$, Xuan-Nam Tran${}^*$, and  Thien Van Luong${}^\ddag$\correspondingauthor} 

\address{
\dag AIoT Lab, Faculty of Computer Science, Phenikaa University, Hanoi, Vietnam\\
${}^*$Advanced Wireless Communications
Group, Le Quy Don Technical University, Hanoi, Vietnam\\
% 6--3--1, Niijuku, Katsushika-ku, Tokyo 125-8585, Japan, \\
\ddag Business AI Lab, Faculty of DS\&AI, National Economics University, Hanoi, Vietnam\\
% 4--1 Gakuendai, Miyashiro-machi, Minamisaitama-gun, Saitama, 345-8501 Japan \\[5pt]
{Email: \email{thienlv@neu.edu.vn}}
}

\maketitle

%%% Notice for ORCiD:
% Do not destroy the format, e.g., do not remove/replace the colons, spaces, or commas.
% Though do not hesitate to add another record following the form.
% \orcid{First Author: 0000-0002-0999-1950}

\abstract
This paper proposes a novel intrusion detection method for unmanned aerial vehicles (UAV) in the presence of recent actual UAV intrusion dataset. In particular, in the first stage of our method, we design an autoencoder architecture for effectively extracting important features, which are then fed into various machine learning models in the second stage for detecting and classifying attack types. To the best of our knowledge, this is the first attempt to propose such the autoencoder-based machine learning intrusion detection method for UAVs using actual dataset, while most of existing works only consider either simulated datasets or datasets irrelevant to UAV communications. Our experiment results show that the proposed method outperforms the baselines such as feature selection schemes in both binary and multi-class classification tasks.
\endabstract
\vspace{-0.2in}

\section{Introduction}
\vspace{-0.1in}
Drones are aircraft or submarines that are controlled remotely without a human operator, and they are often called unmanned aerial vehicles (UAVs) \cite{gupta2016uav}. With their low cost, flexibility, and ease of deployment, flying technologies have been becoming increasingly attractive for unmanned missions. These vehicles can perform tasks such as surveillance, crowd control, and wireless coverage \cite{gupta2016uav}. In this context,  developing an intrusion detection system (IDS) to ensure safety  for UAVs from attacks is really necessary. 

To the best of the author's knowledge, there have been no studies, which utilizes autoencoder to improve the efficiency of IDS for UAVs in the presence of actual UAV intrusion dataset. Note that the intrusion detection systems for UAVs can use either cyber data or physical data for detecting attacks. Most of existing works in UAV intrusion detection rely either on the simulated datasets or irrelevant datasets (which are not for UAVs), while the actual datasets have been overlooked. Recently, a combination of actual cyber and physical dataset \cite{hassler2024intrusion} has been proved to be more effective in detecting cyber attacks of UAVs than using either of them. Therefore, our current work will focus on developing a robust intrusion detection method for UAVs in the presence of the real UAV intrusion dataset \cite{hassler2024intrusion} rather than the simulated datasets or the irrelevant datasets.

\vspace{-0.2in}
\section{Related Works}
\label{sec:related_works}
 \vspace{-0.1in}

\subsection{Related Works in Intrusion Detection for UAVs}
\vspace{-0.1in}
As mentioned early, most of research works in UAV intrusion detection utilize either the simulated datasets or the datasets irrelevant to UAVs. For example, in \cite{han2023}, an IDS for UAV that uses a hierarchical LSTM model to secure packet information was proposed, where the CICIDS-2017 dataset \cite{Sharafaldin2018TowardGA} was used for to demonstrate its ability of effectively detecting anomalies in UAV communications. Also relying on CICIDS-2017, in \cite{bouhamed2021},  a reinforcement Q-learning-based lightweight IDS was developed for detecting cyber attacks in  UAVs. In addition, \cite{kou2022} combined a deep autoencoder and a convolutional neural network (CNN) for detecting malicious attacks to drones under software-defined network environments, using the virtualized InSDN dataset \cite{Elsayed2020InSDN}. In the context of UAV-delivered systems, in \cite{ferrag2019}, a variety of machine learning models were developed in combination with 
the blockchain technique for detecting attacks for reducing latency, using the CSE-CIC-IDS2018 dataset  \cite{Sharafaldin2018TowardGA}. 

As such, all of the aforementioned research works are based on either simulated datasets such as InSDN \cite{Elsayed2020InSDN} or irrelevant datasets such as CSE-CIC-IDS2018 and  CICIDS-2017 \cite{Sharafaldin2018TowardGA}. Recently, a actual dataset for UAV intrusion detection was proposed in \cite{hassler2024intrusion}, which consists of both cyber and physical features. Note that the cyber features are related to communication protocols such as packet size, MAC/IP addresses, while physical features are about physical information of UAVs such as its speed and directions. Also in \cite{hassler2024intrusion}, various machine learning-based detection methods were considered, which are fed by a subset of important features selected based on the Shapley additive explanations (SHAP) analysis. However, such the feature selection schemes may not be optimal in extracting most important features. This motivates us to consider a more advanced method which relies on an autoencoder  for better feature extraction, as will be presented in Section~\ref{sec:proposed_method}. 
\vspace{-0.1in}
\subsection{Autoencoder-based Intrusion Detection for UAVs}
\vspace{-0.1in}
We now review the recent advances in the autoencoder-based IDS for UAVs. Unlike feature selection schemes \cite{Ngo2024Machine}, which simply choose a subset of available features based on some pre-defined criteria \cite{Ngo2024Machine}, the autoencoder-based feature extraction method aims to compress a high-dimensional data into a low-dimensional one by training the autoencoder using a reconstruction loss. 

There are a few of research works that applied autoencoder for extracting helpful features in UAV intrusion detection. For example, in \cite{Park2021Unsupervised}, an autoencoder-based method using the ReLU activation was developed for both fault detection and attack detection in UAVs, where actual physical ALFA \cite{Azarakhsh2021ALFA} and UAV attack datasets \cite{Whelan2020Novelty} were used. Herein, the ALFA dataset has two attack types, namely, GPS spoofing and DoS. In \cite{kou2022}, a deep autoencoder was proposed to reduce data dimensionality and improve training efficiency of IDS, where a CNN classifier was used for classifying attack types based on features extracted by autoencoder in the presence of the virtual InSDN dataset \cite{Elsayed2020InSDN}.

As mentioned above, the actual intrusion datasets for UAVs are not popular, while the unique actual dataset for both cyber and physical features has been published very recently in \cite{hassler2024intrusion}. Thus, the application of autoencoder and machine learning models to such actual dataset has been overlooked in the literature. This work aims to fill this research gap by proposing a novel intrusion detection method for UAVs that employs an autoencoder architecture for effectively extracting important information from the original data as well as for reducing data dimensionality, where the actual cyber dataset in \cite{hassler2024intrusion} is used. Then, various machine learning models such as Random Forest (RF), Support Vector Machine (SVM), K-Nearest Neighbors (KNN), Decision Tree (DT), Multi-Layer Perceptron (MLP) are adopted to process the low-dimensional data extracted by autoencoder for reliably detecting cyber attacks in UAVs.

\vspace{-0.15in}
\section{Proposed Method}\label{sec:proposed_method}

\begin{figure}[ht]
    \centering
    \includegraphics[width=1\columnwidth]{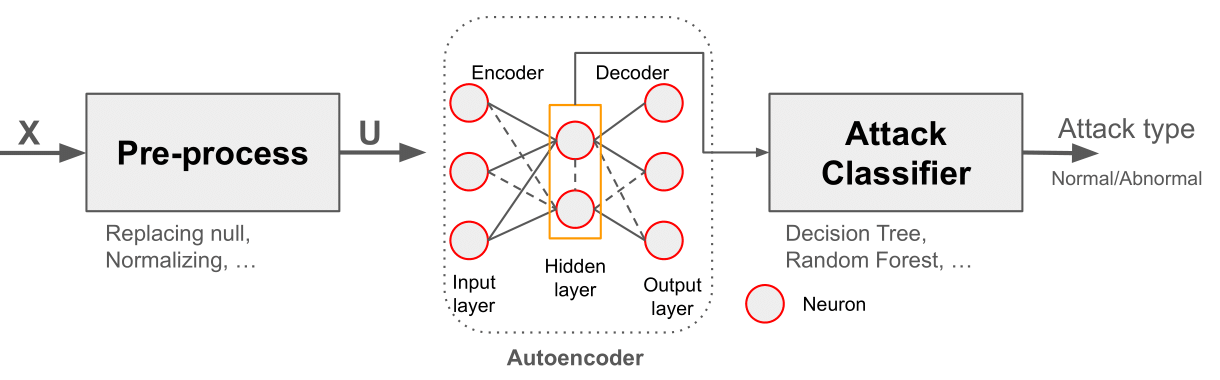}
    \caption{Block diagram of the proposed autoencoder-based intrusion detection system for UAV communications.}
    \label{fig:pipeline-overview}
\end{figure}
\vspace{-0.1in}

Our IDS, as shown in Figure~\ref{fig:pipeline-overview}, comprises three main components, namely, data pre-processing, feature extraction, and attack classification. The workflow begins with pre-processing the raw data \textbf{X}, such as handling null values and data normalization. Then, an autoencoder (AE) is introduced for feature reduction, which is capable of capturing intricate patterns in network traffic data, providing a more effective feature reduction than traditional methods. Finally, the AE-extracted data is fed to the attack classifier, which employs machine learning models to classify the network traffic into Normal or Abnormal, and further identify specific attack types. 

\begin{table}[ht]
\centering
\caption{Autoencoder architecture configuration}
\label{tab:autoencoder-config}
\begin{tabular}{|l|c|c|c|}
\hline
\textbf{Layer} & \textbf{Dimension} & \textbf{Activation} & \textbf{Parameters} \\
\hline
Input & $M$ & - & 0 \\
\hline
\multicolumn{4}{|c|}{\textbf{Encoder}} \\
\hline
Dense 1 & 40 & tanh & $40M+40$ \\
Dense 2 & 20 & tanh & 820 \\
Dense 3  & $N$ & linear & $21N$ \\
\hline
\multicolumn{4}{|c|}{\textbf{Decoder}} \\
\hline
Dense 4 & 20 & tanh & $20N+20$ \\
Dense 5 & 40 & tanh & 840 \\
Dense 6  & $M$ & linear & $41M$ \\
\hline
\multicolumn{4}{|l|}{\textbf{Total model parameters:} $41N+81M+1720$} \\
\hline
\end{tabular}
\end{table}

% \begin{table*}[ht]
%     \centering
%     \caption{Features selected in two existing feature selection methods in \cite{hassler2024intrusion}}\label{tab:feature-selected}
%     \resizebox{1\textwidth}{!}{\begin{tabular}{|c|c|c|}
%     \hline 
%     \multirow{1}{*}{Selected features} & \multicolumn{1}{c|}{SVM-SHAP} & \multicolumn{1}{c|}{FNN-SHAP}\tabularnewline
%     \hline 
%     4 & wlan.fc.type, wlan.duration, ip.hdr\_len, ip.dst & wlan.duration, ip.hdr\_len, ip.dst, frame.len\tabularnewline
%     \hline 
%     \multirow{2}{*}{8} & wlan.fc.type, wlan.duration, ip.hdr\_len, ip.dst, & wlan.duration, ip.hdr\_len, ip.dst, frame.len, ip.ttl\tabularnewline
%      & frame.len, ip.proto, frame.protocols, wlan.seq & frame.protocols, time\_since\_last\_packet, udp.dstport
%     \tabularnewline
%     \hline 
%     \end{tabular}}

% \end{table*}

\begin{table*}[ht]
\centering \caption{Performance of different machine learning models using the proposed autoencoder for extracting 4 features}
\label{tab:4-ml} \linespread{0.0} %

% Preview source code for paragraph 0

\begin{tabular}{|c|c|c|c|c|c|c|c|c|}
\hline 
\multirow{2}{*}{Models} & \multicolumn{4}{c|}{Binary classification} & \multicolumn{4}{c|}{Multi-class classification}\tabularnewline
\cline{2-9} \cline{3-9} \cline{4-9} \cline{5-9} \cline{6-9} \cline{7-9} \cline{8-9} \cline{9-9} 
 & Precision & Recall & F1-score & Accuracy & Precision & Recall & F1-score & Accuracy\tabularnewline
\hline 
DT & \textcolor{red}{\textbf{91.63}} & \textcolor{red}{\textbf{88.74}} & \textcolor{red}{\textbf{89.67}} & \textcolor{red}{\textbf{88.74}} & 79.15 & 79.66 & 79.4 & 72.48 \\ \hline
RF & 90.52 & 87.34 & 88.06 & 87.34 & 81.01 & 80.47 & 80.74 & 73.01 \\ \hline
KNN & 77.55 & 80.13 & 79.62 & 80.13 & 80.57 & 81.21 & 80.89 &\textcolor{red}{\textbf{74.55}} \\ \hline
MLP & 90.53 & 87.56 & 88.74 & 87.56 & \textcolor{red}{\textbf{83.18}} & \textcolor{red}{\textbf{81.29}} & \textcolor{red}{\textbf{82.22}} & 74.02 \\ \hline
SVM & 84.26 & 85.13 & 84.37 & 85.13 & 48.69 & 50.27 & 49.52 & 50.27 \\ \hline
\end{tabular}

\end{table*}

\begin{table*}[ht]
\centering \caption{Performance of different machine learning models using the proposed autoencoder for extracting 8 features}
\label{tab:8-ml} \linespread{1.0} %

% Preview source code for paragraph 0

\begin{tabular}{|c|c|c|c|c|c|c|c|c|}
\hline 
\multirow{2}{*}{Models} & \multicolumn{4}{c|}{Binary classification} & \multicolumn{4}{c|}{Multi-class classification}\tabularnewline
\cline{2-9} \cline{3-9} \cline{4-9} \cline{5-9} \cline{6-9} \cline{7-9} \cline{8-9} \cline{9-9} 
 & Precision & Recall & F1-score & Accuracy & Precision & Recall & F1-score & Accuracy\tabularnewline
\hline 
DT & \textcolor{red}{\textbf{94.87}} & \textcolor{red}{\textbf{94.54}} & \textcolor{red}{\textbf{94.21}} & \textcolor{red}{\textbf{94.54}} & 79.66 & 80.24 & 79.95 & 73.19 \\ \hline
RF & 92.31 & 91.45 & 91.23 & 91.45 & 81.79 & 80.77 & 81.28 & 73.34 \\ \hline
KNN & 88.27 & 88.61 & 88.78 & 88.61 & 81.45 & 82.12 & 81.78 & \textcolor{red}{\textbf{75.72}} \\ \hline
MLP & 93.22 & 91.19 & 92.54 & 91.19 & \textcolor{red}{\textbf{85.99}} & \textcolor{red}{\textbf{82.26}} & \textcolor{red}{\textbf{84.09}} & 75.34 \\ \hline
SVM & 83.48 & 84.19 & 84.33 & 84.19 & 60.73 & 59.41 & 60.42 & 59.41 \\ \hline
\end{tabular}

\end{table*}

\vspace{-0.1in}

\subsection{Autoencoder-based Feature Extraction}
\vspace{-0.1in}
\begin{figure}[ht]
    \centering
    \includegraphics[width=0.95\linewidth]{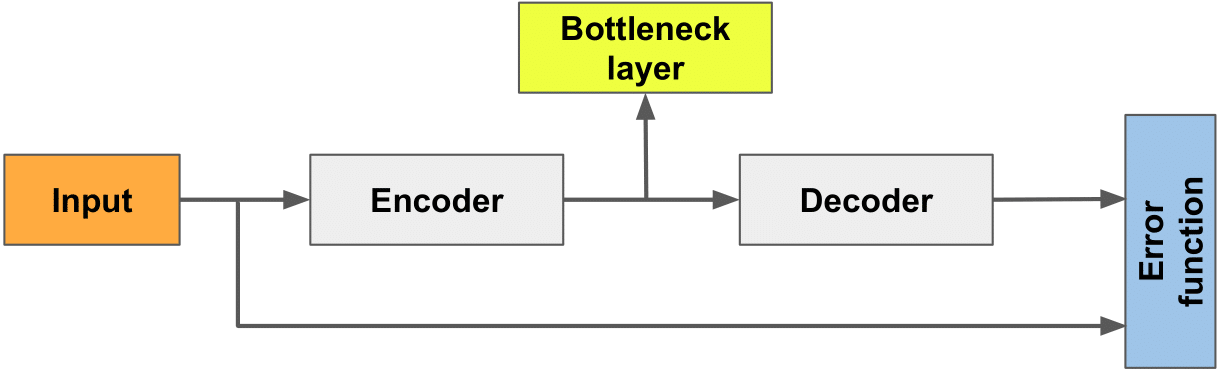}
    \caption{Autoencoder representation.}
    \label{fig:ae-representation}
\end{figure}

The AE, depicted in Figure~\ref{fig:pipeline-overview}, which  transforms  feature vectors into abstract representations as shown in Figure~\ref{fig:ae-representation}, is an unsupervised neural network with input, hidden, and output layers. The encoding process maps the input vector $\mathbf{x} \in \mathbb{R}^{M}$ ($M$ is the number of input features) to a low-dimensional representation $\mathbf{h}\in \mathbb{R}^N$ using the transformation: $\mathbf{h} = g_{\theta_1}(\mathbf{x}) = \mathbf{W}_{1q}\sigma\left(...\sigma\left(\mathbf{W}_{11}\mathbf{x}+\mathbf{b}_{11}\right)...\right)+\mathbf{b}_{1q},$ where $\mathbf{W}_{1i}$ is the weight matrix and $\boldsymbol{b}_{1i}$ is the bias vector for the the $i$-th encoding dense layer, for $i=1,2,...,q$ and $q$ is the number of dense layers of the encoder and decoder. The decoding process reconstructs the input vector $\mathbf{x}$ from $\mathbf{h}$ to $\mathbf{y} \in \mathbb{R}^{M}$ using the transformation: $\mathbf{y} = g_{\theta_2}(\mathbf{h}) = \mathbf{W}_{2q}\sigma\left(...\sigma\left(\mathbf{W}_{21}\mathbf{h}+\mathbf{b}_{21}\right)...\right)+\mathbf{b}_{2q},$
where $\mathbf{W}_{2i}$ is the weight matrix and $\boldsymbol{b}_{2i}$ is the bias vector for the $i$-th dense layer of the decoder. The activation function $\sigma$ used in both encoding and decoding layers is the hyperbolic tangent (tanh) function, defined as: $
\sigma(x) = \tanh(x) = \frac{e^x - e^{-x}}{e^x + e^{-x}}.$ The detailed AE design in this study is shown in Table~\ref{tab:autoencoder-config}, where the encoder and decoder have $q=3$ dense layers and the output layers of both encoder and decoder are linear layers.

The objective is to minimize the reconstruction error given by the mean squared error between $\mathbf{x}$ and $\mathbf{y}$:
\begin{equation}
L(\theta_1, \theta_2) = \frac{1}{2T} \sum_{j=1}^T \|\mathbf{x}^{(j)} - \mathbf{y}^{(j)}\|^2,\label{eq:loss-function}
\end{equation}
where $T$ is the number of training samples, $\theta_1, \theta_2$ represent the weight matrices and biases, i.e., $\{\mathbf{W}_{1i}, \boldsymbol{b}_{1i}\}_{i=1}^{q}$ and $\{\mathbf{W}_{2i}, \boldsymbol{b}_{2i}\}_{i=1}^{q}$ of the encoder and decoder, respectively.

\vspace{-0.1in}
\subsection{Machine Learning-based Intrusion Detection}
\vspace{-0.1in}
Machine learning-based IDS classifies network traffic and detects anomalies using algorithms such as Random Forest (RF), Support Vector Machine (SVM), K-nearest neighbors (KNN), Decision Tree (DT), and Multi-layer Perception (MLP) \cite{Ngo2024Machine}. Key pre-processing steps include label encoding and feature scaling. 
% Various machine learning models are implemented to classify intrusions, enhancing detection accuracy and mitigating overfitting.

\begin{figure}[ht]
    \centering
    \includegraphics[width=0.9\linewidth]{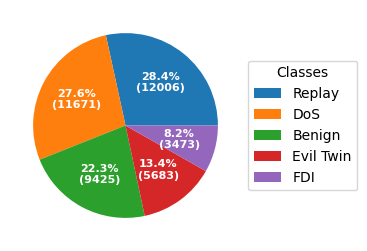}
    \caption{Proportions of 5 classes of actual cyber data \cite{hassler2024intrusion}.}
    \label{fig:enter-label}
\end{figure}

\vspace{-0.2in}

\section{Experimental Results and Discussion}\label{sec:results}
\vspace{-0.1in}
\subsection{Overview of Dataset}\label{subsec: overview-data}
\vspace{-0.1in}
We evaluate our method using the actual UAV intrusion dataset from \cite{hassler2024intrusion}, which contains around 42,000 records, where there are a Benign class and 4 attack classes, namely, De-Authentication (DoS), Replay, Evil Twin, and False Data Injection (FDI). Figure~\ref{fig:enter-label} shows the class distribution. Note that non-essential features such as frame.number, wlan.bssid, and timestamp\_c were excluded. Thus, the number of remaining features are $M=54$ (see Table~\ref{tab:autoencoder-config}).

\vspace{-0.1in}
\subsection{Implementation Setting}\label{subsec:settings}

\vspace{-0.1in}

\begin{table*}[ht]
\centering \caption{Comparison between the proposed autoencoder and existing feature selection methods with 4 extracted features}
\label{tab:4} \linespread{1.0} %

% Preview source code for paragraph 0

\begin{tabular}{|c|c|c|c|c|c|c|c|c|}
\hline 
\multirow{2}{*}{Models} & \multicolumn{4}{c|}{Binary classification} & \multicolumn{4}{c|}{Multi-class classification}\tabularnewline
\cline{2-9} \cline{3-9} \cline{4-9} \cline{5-9} \cline{6-9} \cline{7-9} \cline{8-9} \cline{9-9} 
 & Precision & Recall & F1-score & Accuracy & Precision & Recall & F1 & Accuracy\tabularnewline
\hline 
SVM-SHAP & \textcolor{red}{\textbf{97.48}} & 85.71 & 91.23 & 86.55 & 75.66 & 67.55 & 71.38 & 66.98 \tabularnewline
\hline
FNN-SHAP & 96.87 & 91.11 & 94.44 & 91.27 & 78.32 & 79.30 & 78.81 & 71.64  \tabularnewline
\hline
Proposed autoencoder & 97.18 & \textcolor{red}{\textbf{94.73}} & \textcolor{red}{\textbf{96.27}} & \textcolor{red}{\textbf{93.85}} & \textcolor{red}{\textbf{83.18}} & \textcolor{red}{\textbf{81.29}} & \textcolor{red}{\textbf{82.22}} & \textcolor{red}{\textbf{74.02}} \\ \hline
\end{tabular}

\end{table*}

\begin{table*}[ht]
\centering \caption{Comparison between the proposed autoencoder and existing feature selection methods with 8 extracted features}
\label{tab:8} \linespread{1.0} %

\begin{tabular}{|c|c|c|c|c|c|c|c|c|}
\hline 
\multirow{2}{*}{Models} & \multicolumn{4}{c|}{Binary classification} & \multicolumn{4}{c|}{Multi-class classification}\tabularnewline
\cline{2-9} \cline{3-9} \cline{4-9} \cline{5-9} \cline{6-9} \cline{7-9} \cline{8-9} \cline{9-9} 
 & Precision & Recall & F1-score & Accuracy & Precision & Recall & F1 & Accuracy\tabularnewline
\hline 
SVM-SHAP & \textcolor{red}{\textbf{98.17}} & 94.12 & 95.02 & 93.47 & 79.61 & 80.27 & 80.15 & 74.71 \tabularnewline
\hline
FNN-SHAP & 94.11 & 95.23 & 96.14 & \textcolor{red}{\textbf{94.96}} & 81.78 & 80.02 & 80.89 & 72.37 \tabularnewline
\hline
Proposed autoencoder & 96.78 & \textcolor{red}{\textbf{95.37}} & \textcolor{red}{\textbf{96.72}} & 94.52 & \textcolor{red}{\textbf{85.99}} & \textcolor{red}{\textbf{82.26}} & \textcolor{red}{\textbf{84.09}} & \textcolor{red}{\textbf{75.34 }} \tabularnewline
\hline 
\end{tabular}

\end{table*}

\textbf{Evaluation Metrics:} The evaluation metrics considered in this study include Precision, Recall, F1-score, and Accuracy, whose detailed definitions can be found in \cite{Ngo2024Machine}.

\textbf{Autoencoder Architecture:} The autoencoder consists of an encoder and a decoder, each with three dense layers, with the encoding dimension \(N\) serving as the bottleneck and the input dimension \(M\) representing the number of features in the dataset. The model is compiled with the Adam optimizer and mean squared error loss given \eqref{eq:loss-function}.

% The autoencoder consists of an encoder and a decoder, each with three dense layers, with the encoding dimension \(N\) serving as the bottleneck and the input dimension \(M\) representing the number of features in the dataset. This structure, shown in Table~\ref{tab:autoencoder-config}, enables dimensionality reduction and data reconstruction. The model uses the Adam optimizer and mean squared error loss as defined in Equation~\ref{eq:loss-function}.

\textbf{Evaluation Procedure:} We evaluate the proposed autoencoder-based method using various machine learning models, from which we select the best models for binary and multi-classification tasks for comparing with the baselines, namely, SVM-SHAP and FNN-SHAP feature selection methods \cite{hassler2024intrusion}. For this, we consider $N=4, 8$ extracted features for performance evaluation of all schemes.

\vspace{-0.15in}

\subsection{Performance Comparison and Discussion}
\vspace{-0.1in}

Table~\ref{tab:4-ml} and Table~\ref{tab:8-ml} illustrate the  performance of the proposed autoencoder method with different machine learning models, for both binary and multi-class classification tasks, in the presence of 4 and 8 extracted features. It is worth noting from these two table that increasing the number of extracted features helps improve the detection performance, particular for binary classification. For example, the F1-score and accuracy of the best binary classifier, i.e., DT, increases from 89.67\% and 88.74\% to 94.21\% and 94.64\%, respectively. However, in multi-class classification, MLP performs the best among classifiers, in terms of Precision, Recall, and F1-score metrics. Therefore, we will employ DT and MLP classifiers for comparing with  existing feature selection methods \cite{hassler2024intrusion} in the following.

In Table~\ref{tab:4} and Table~\ref{tab:8}, we compare the performance between the proposed autoencoder-based method and existing feature selection methods, namely, SVM-SHAP and FNN-SHAP \cite{hassler2024intrusion}. In all schemes, as mentioned in Table~\ref{tab:4-ml} and Table~\ref{tab:8-ml}, DT and MLP classifiers are used for binary and multi-class classification, respectively. It is shown via Table~\ref{tab:4} and Table~\ref{tab:8} that our method outperforms the baselines in a majority of metrics, especially for multi-class classification. For example, in Table~\ref{tab:8}, in multi-class classification, our method achieves a F1-score of 84.09\%, which is much higher than that of SVM-SHAP and FNN-SHAP with 80.15\% and 80.89\%, respectively. This confirms the effectiveness of the proposed autoencoder-based learning method for UAV intrusion detection.
\vspace{-0.15in}

\section{Conclusions}
\label{sec:conclusion}
\vspace{-0.15in}
We proposed an effective autoencoder-based machine learning intrusion detection method for UAV communications for the first time, in the presence of the actual cyber dataset. Our method relies on an autoencoder neural network to extract important features from the original data, which are then fed to machine learning models for classifying attack types. We evaluated our proposed method under both binary and multi-class classification tasks, where experiment results showed that using autoencoder-based feature extraction, Detection Tree is the best binary classifier, while MLP is the best multi-class classifier. More importantly, the proposed method outperforms the existing feature selection schemes in terms of various performance metrics, particularly in multi-class classification tasks.

\vspace{-0.15in}

\section*{Acknowledgments}
\vspace{-0.1in}
This research was supported by Vietnam National Foundation for Science and Technology Development (NAFOSTED) under grant number 102.02-2021.56.
\vspace{-0.15in}
% \printbibliography
\bibliographystyle{plain}
% \bibliography{refs.bib}
% \bibliographystyle{plainnat}
\bibliography{refs}

\end{document}